%% file: paper.tex
\documentclass[runningheads]{llncs}
\usepackage{graphicx}
\usepackage{hyperref}
\usepackage{tabularx}
\usepackage[export]{adjustbox}
\begin{document}

\title{Brilla AI: AI Contestant for the National Science and Maths Quiz}



\author{George Boateng\inst{1,2} \and
Jonathan Abrefah Mensah\inst{1}\and
Kevin Takyi Yeboah \inst{1}\and
William Edor \inst{1}\and
Andrew Kojo Mensah-Onumah \inst{1}\and
Naafi Dasana Ibrahim \inst{1}\and
Nana Sam Yeboah \inst{1} \\}
\authorrunning{Boateng et al.}

\institute{Kwame AI Inc., U.S.\\ 
\and
ETH Zurich, Switzerland}


\maketitle

\begin{abstract}
The African continent lacks enough qualified teachers which hampers the provision of adequate learning support. An AI could potentially augment the efforts of the limited number of teachers, leading to better learning outcomes. Towards that end, this work describes and evaluates the first key output for the NSMQ AI Grand Challenge, which proposes a robust, real-world benchmark for such an AI: “Build an AI to compete live in Ghana’s National Science and Maths Quiz (NSMQ) competition and win — performing better than the best contestants in all rounds and stages of the competition”. The NSMQ is an annual live science and mathematics competition for senior secondary school students in Ghana in which 3 teams of 2 students compete by answering questions across biology, chemistry, physics, and math in 5 rounds over 5 progressive stages until a winning team is crowned for that year. In this work, we built Brilla AI, an AI contestant that we deployed to unofficially compete remotely and live in the Riddles round of the 2023 NSMQ Grand Finale, the first of its kind in the 30-year history of the competition. Brilla AI is currently available as a web app that livestreams the Riddles round of the contest, and runs 4 machine learning systems:  (1) speech to text (2) question extraction (3) question answering and (4) text to speech that work together in real-time to quickly and accurately provide an answer, and then say it with a Ghanaian accent. In its debut, our AI answered one of the 4 riddles ahead of the 3 human contesting teams, unofficially placing second (tied). Improvements and extensions of this AI could potentially be deployed to offer science tutoring to students and eventually enable millions across Africa to have one-on-one learning interactions, democratizing science education.


\end{abstract}



\input{content.tex}


\section{Acknowledgments}
We are grateful to all the volunteers that contributed to this project. We are also grateful to Isaac Sesi for donating his voice which we used to train the TTS model for the voice of our AI, and our project advisors for their feedback and support throughout this work: Professor Elsie Effah Kaufmann, the NSMQ Quiz Mistress, and Timothy Kotin, a member of the 2006 NSMQ winning team.

\bibliographystyle{splncs04}
\bibliography{refs}


\end{document}

%% file: content.tex
\section{Introduction}
According to UNESCO, only 65\% of primary school teachers in Sub-Saharan Africa possessed the necessary minimum qualifications \cite{UNESCO2021}. Moreover, the average student-teacher ratio at the primary education level in Sub-Saharan Africa stood at 38:1 in 2019, a figure significantly higher than the ratio of 13.5:1 observed in Europe \cite{Eurostat2021}. Consequently, the region requires an additional 15 million teachers by 2030 to meet education objectives, a formidable and costly challenge \cite{UNESCO2021b}. The scarcity of adequately qualified educators in Africa undermines the provision of effective learning support for students. Introducing an Artificial Intelligence (AI) teaching assistant for educators holds promise in augmenting the efforts of the limited teaching workforce, facilitating tasks such as one-on-one tutoring and answering student queries. However, the absence of a robust benchmark tailored to real-world scenarios and the African context poses a significant obstacle to evaluating the effectiveness of such AI solutions.

Motivated by this need, Boateng et al. proposed a grand challenge in education — \textbf{NSMQ AI Grand Challenge} — which is a robust, real-world challenge in education for such an AI: “\textit{Build an AI to compete in Ghana’s National Science and Maths Quiz Ghana (NSMQ) competition and win — performing better than the best contestants in all rounds and stages of the competition}” \cite{boateng2023nsmq}. In that work, they detailed the motivation for the challenge and key technical challenges that must be addressed for an AI to win the NSMQ. We created an open-source project \footnote{Brilla AI Open-Source Project: \url{https://github.com/nsmq-ai/nsmqai}} to conquer this grand challenge and built Brilla AI as the first key output which extends our prior work \cite{boateng2023nsmqai}. Brilla AI is an AI contestant that we deployed to unofficially compete remotely and live in the Riddles round of the 2023 NSMQ Grand Finale, the first of its kind in the 30-year history of the competition.

The NSMQ is an exciting, annual live science and mathematics quiz competition for senior secondary school students in Ghana in which 3 teams of 2 students compete by answering questions across biology, chemistry, physics, and math in 5 rounds over 5 progressive stages until a winning team is crowned for that year \cite{NSMQ}. The competition has been run for 30 years and poses interesting technical challenges across speech-to-text, text-to-speech, question-answering, and human-computer interaction. Given the complexity of the challenge, we decided to start with one round, round 5 — Riddles.  This round, the final one, is arguably the most exciting as the competition's winner is generally determined by the performance in the round. In the Riddles round, students answer riddles across Biology, Chemistry, Physics, and Mathematics. Three (3) or more clues are read to the teams that compete against each other to be first to provide an answer (usually a word or a phrase) by ringing their bell. The clues start vague and get more specific. To make it more exciting and encourage educated risk-taking, answering on the 1st clue fetches 5 points, on the 2nd clue — 4 points, and on the 3rd or any clue thereafter, 3 points. There are 4 riddles for each contest with each riddle focusing on one of the 4 subjects. Speed and accuracy are key to winning the Riddles round. An example riddle with clues and the answer is as follows (see a live example here: \footnote{Video Example of Riddle:  \url{https://www.youtube.com/watch?v=kdaxoFjiYJg}}. \textbf{Question:} (1) I am a property of a periodic propagating disturbance. (2) Therefore, I am a property of a wave. (3) I describe a relationship that can exist between particle displacement and wave propagation direction in a mechanical wave. (4) I am only applicable to waves for which displacement is perpendicular to the direction of wave propagation. (5) I am that property of an electromagnetic wave which is demonstrated using a polaroid film. Who am I? \textbf{Answer:} Polarization.

Here are some of the key technical challenges an AI system will need to address. How can it accurately provide real-time transcripts of Ghanaian-accented riddle questions read in English? Speech-to-text systems tend to be trained using data from the West and generally do not work well for African-accented speech \cite{olatunji2023}. How can it infer the start and end of each riddle and extract only the clues while discarding all statements (e.g., the quiz mistress pausing the reading of the clues after the ring of a bell to listen to an answer from a school, instructions that are read at the start of the round, etc.)?  How does the AI know the optimal time to attempt to provide an answer optimizing for speed and accuracy? Providing an answer too early might result in a wrong answer given that earlier clues tend to be vague. Waiting too long could result in an answer being provided by one of the contesting teams. How can the AI say its answer with a Ghanaian accent? Similar to STT systems, several text-to-speech systems tend to be trained using data from the West and generally do not speak with African accents. How do we ensure all these work seamlessly in real time without any noticeable latency? 

Similar work on grand challenges that entail question answering include the DeepQA project in which Watson won \textit{Jeopardy!} in 2011 \cite{watson}, and the International Math Olympiad (IMO) Grand Challenge \cite{imo} which has an education focus. The Brilla AI project has some similarities with Jeopardy! given they are both live quiz shows, that entail question answering. It differs, given its science education focus making it arguably more challenging, and provides unique sets of technical challenges discussed previously and also by Boateng et al. \cite{boateng2023nsmq}. Despite the similarity with the IMO Grand Challenge as both focus on STEM education, it differs by having a scope of both science and math education, and consists of a live competition (IMO is only written). The most important way Brilla AI differs from these 2 challenges though is the African context focus, a rarity in the literature for grand challenges which makes this work the first of its kind. 

Our contribution is the first end-to-end, real-time AI system deployed as an unofficial, AI contestant together with 3 human competing teams for the 2023 NSMQ Grand Finale (the first of its kind in the 30-year history of the competition) that (1) transcribes Ghanaian-accented speech of scientific riddles, (2) extracts relevant portions of the riddles (clues) by inferring the start and end of each riddle and segmenting the clues, (3) provides an answer to the riddle, and (4) then says it with a Ghanaian accent.

\section{Background}
The task of question answering has two primary paradigms: Extractive QA and Generative QA. Extractive QA involves machine reading, wherein a segment of text within a larger body (referred to as the context) is selected to directly answer a question. The context may be provided directly, as exemplified by the task outlined in the SQuAD dataset. Alternatively, relevant passages can be retrieved from diverse documents through a process known as retrieval. Chen et al. employed this approach in their DrQA system, utilizing TF-IDF for retrieval and LSTM for answer extraction \cite{chen2017}. Presently, BERT-based models are the current state-of-the-art for reader implementation \cite{devlin2018}. In contrast, Generative QA employs generative models such as T5 \cite{roberts2020} to produce answers given a question as input. These models can generate responses based on various contexts provided alongside the question, a technique termed retrieval augmented generation (RAG) \cite{lewis2020}. We use both extractive QA and generative QA in this work and leave RAG for future work.

\section{Brilla AI System}
\begin{figure}
\centering
\includegraphics[width=\linewidth]{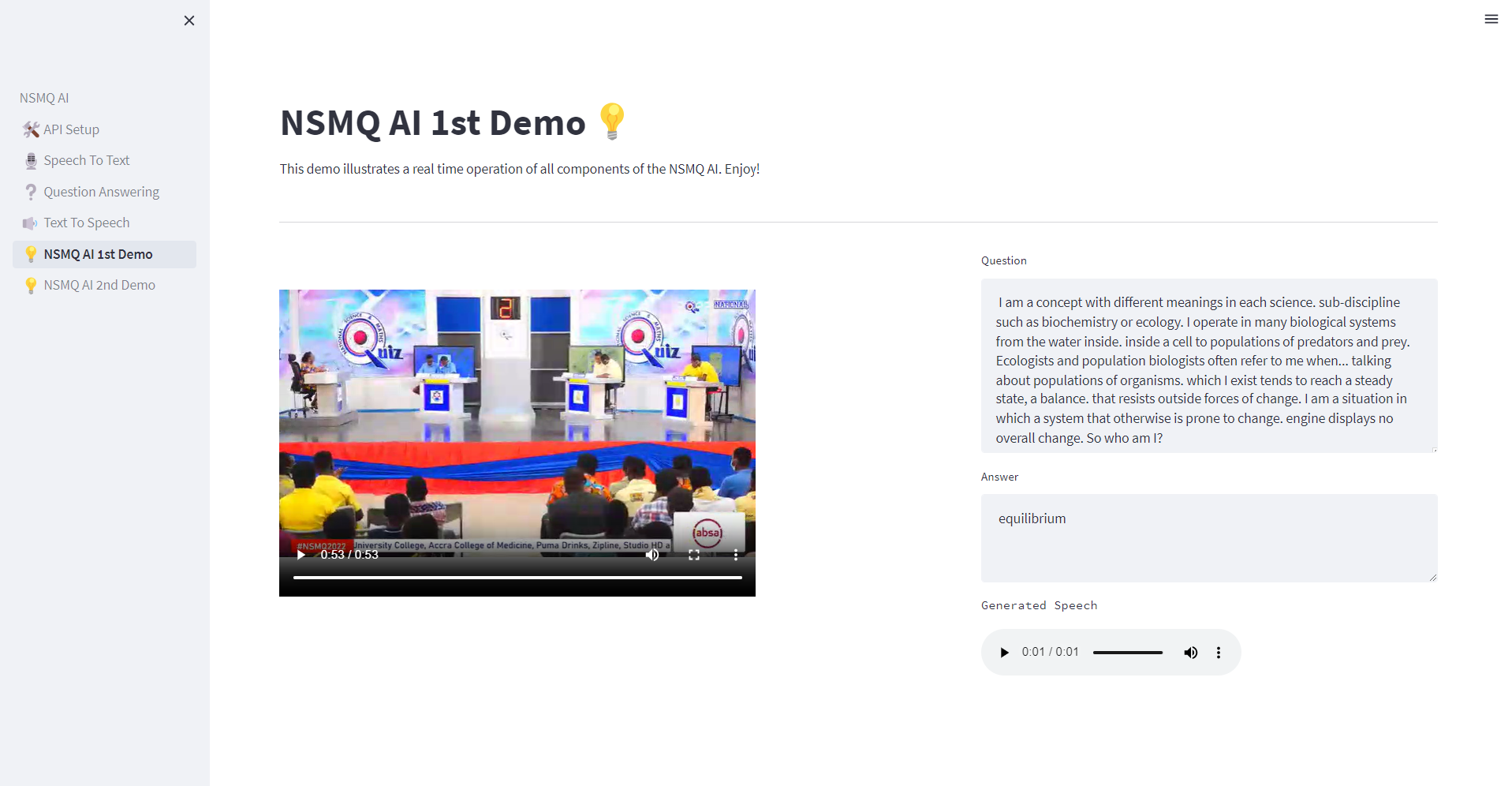}
\caption{Screenshot of Brilla AI web app} 
\label{fig:webapp}
\end{figure}

\begin{figure}[ht]
\centering
\includegraphics[width=\linewidth]{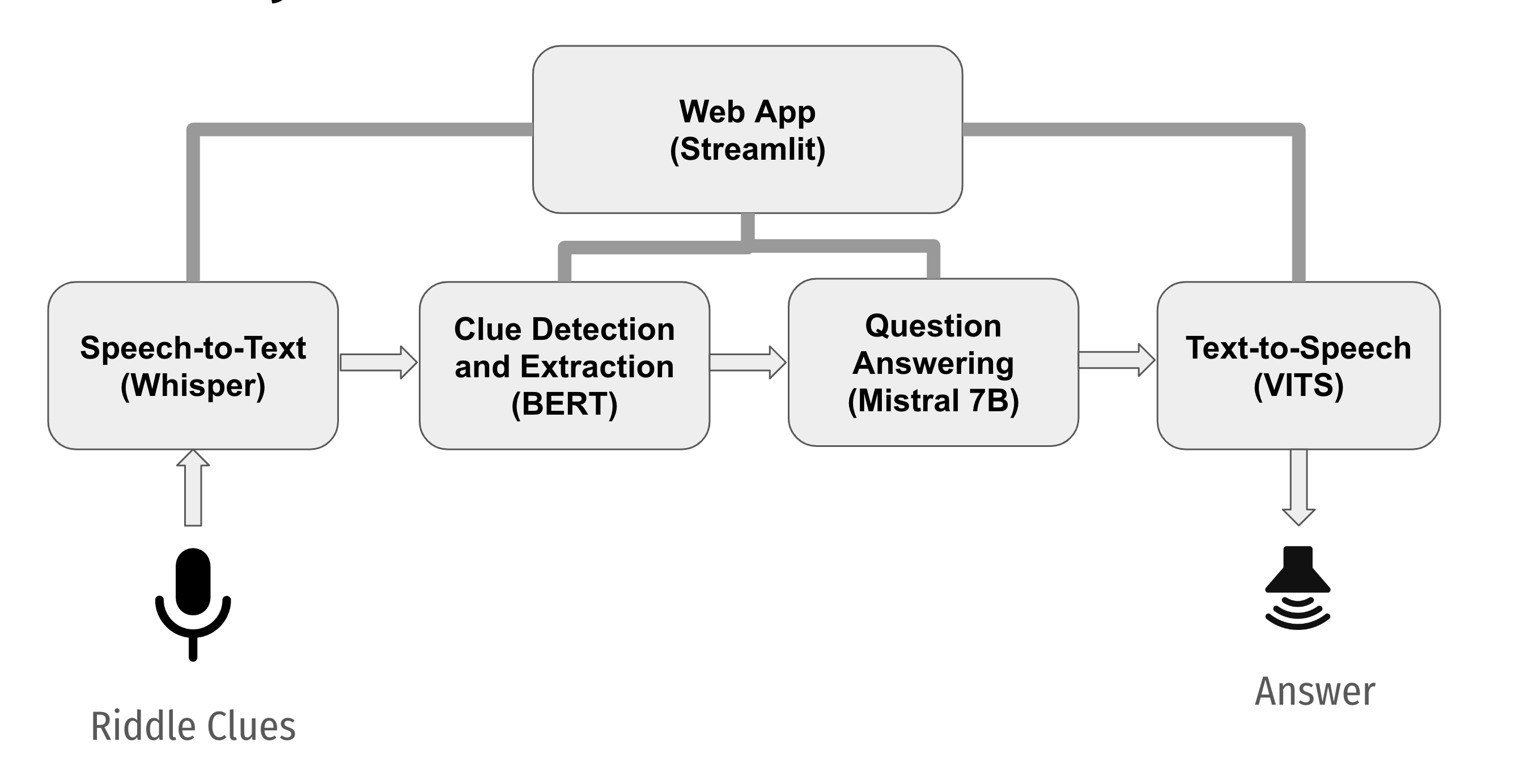}
\caption{Brilla AI System} 
\label{fig:brilla_ai}
\end{figure}

Brilla AI is currently available as a web app (built with Streamlit)  (Fig \ref{fig:webapp}) that livestreams the Riddles round of the contest, and runs 4 machine-learning (ML) systems in real-time via FastAPI together running as an AI server on Google Colab: (1) speech-to-text (using Whisper \cite{radford2023}) which transcribes Ghanaian-accented speech of scientific riddles (2) question extraction (using BERT \cite{devlin2018}) which extracts relevant portions of the riddles (clues) by inferring the start and end of each riddle and segmenting the clues (3) question answering (using Mistral \cite{jiang2023}) which provides an answer to the riddle and (4) text-to-speech (using VITS \cite{kim2021}) which says the answer with a Ghanaian accent (Figure \ref{fig:brilla_ai}). We used Streamlit due to its applicability to ML and Data science use cases, and quick development cycles.

Here is an overview of the sequential processing and communication between components in the Brilla AI system. The web app extracts audio from the video and chunks it in 5 seconds using FFmpeg to allow for low-latency processing, and then sends and receives relevant data from the AI server sequentially. First, the audio chunk is sent to the STT API for processing which extracts the transcript, and passes it to the QE service which detects the riddle start and extracts the clues. The clues along with information about whether it is a new riddle are sent to the web app. It displays the transcript and clues and then sends the clues to the QA API which determines whether to provide an answer (details described later). If an answer is generated, it is returned to the web app which then displays it and sends it to the TTS API. It generates an audio of the answer and sends it to the web app which then plays it.  This current sequential operation is a bottleneck within the system which caused some latency issues (discussed later). We are exploring parallelizing operations to remove such bottlenecks from the system.

\begin{figure}[ht]
\centering
\includegraphics[width=\linewidth]{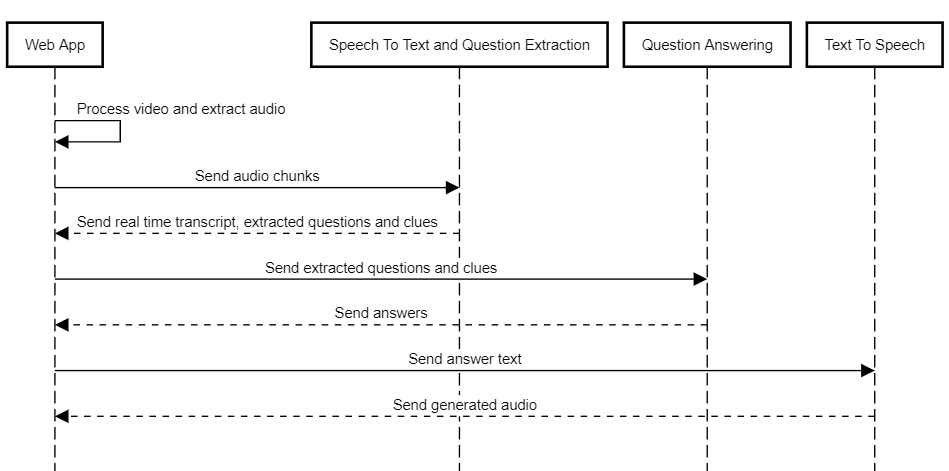}
\caption{Sequential Processing in Brilla AI System} 
\label{fig:sequential_processing}
\end{figure}

The web app consists of 2 modes: demo and live \footnote{Brilla AI Demos: \url{https://youtube.com/playlist?list=PLl-Yq3F_E8VJWFLpsnizorhIz8N-OqUqw&si=ddrl7u0DBZcbNt7H}}. The demo mode showcases the AI functionality outside of the quiz season. Functionality has been added for users to experience the AI using curated text, audio, and video content but also allows them to provide real-time content to test the AI. For example, in the demo mode, the user can provide real-time audio input on the  Speech-to-Text page. The web app processes this audio and sends an API call to the Speech-to-Text API server. The server returns the real-time transcript displayed in the web app to the user. Also, users can provide a YouTube link of any NSMQ quiz video to the web app and see the AI process the content in real time and compete for that video. To demonstrate the AI during active quiz season, the web app utilizes a live mode. In this mode, the user can supply a live stream link from YouTube to the web app and see the AI compete in real-time as well.

\section{Data Collection and Preprocessing}
We curated a dataset of NSMQ contests from 2012-2022 containing videos of the contest and corresponding metadata, text form of riddles questions, and open-source science textbooks. First, we created a Google sheet with information about each contest such as the contest date, names of the competing schools, and the marks they acquired at the end of each contest. We additionally provided information for the riddles round such as which school answered each riddle and at which clue, which was used to calculate the points obtained by the best human contestants for each riddle. We then used that information to automatically extract video clips of the contests. Next, we manually generated annotations for the riddle round containing information about the start and end timestamps of each riddle and clue. These were then used to automatically extract the video and audio clip segments of riddles. 

For the text of riddles, we purchased the digital version of the questions, parsed them, and manually reformatted them into CSVs for usage. Each CSV file had columns “Clue 1” to “Clue 9” for all the clues in per riddle, “Answer” for the ground truth answer, and “Answer 1” to “Answer 4” for alternative ground truth answers, if any, (e.g., hydrogen and h2 as alternative answers). We included additional columns to track information like subject, contest number, and year of the contest. We applied the following preprocessing steps and created riddle-answer mappings: converted all clue texts and answers to lowercase, removed punctuation, fixed whitespace, and removed articles (e.g., “the”, “a” “an”).

We extracted HTML files from the following open-source Science textbooks on OpenStax — High School Physics, College Physics, College Biology, College Chemistry, and College Algebra — segmenting content into chapters, sections, paragraphs, and passages. We have completed the sheet with information about contests (with some contest information missing for some years as they could not be found online). For information about the performance of contestants in the riddle round, we only have complete information for the years 2019 and 2020. Though we have complete data for the text version of the riddle questions, we only used riddles for these 2 years for evaluation so we could compare the performance of our system with human performances. Consequently, we used a total of 316 riddles (156 in 2019 and 160 in 2020). Work is ongoing to annotate the timestamps of each of the clues for all riddles and also to completely parse the textbooks. Hence, we did not have a complete set of video and audio clips of riddles for this work.

\section{Modeling, Experiments, and Evaluation}
\subsection{Speech-to-Text}
The Speech-To-Text (STT) system \footnote{STT System Demo: \url{https://youtu.be/jn4Kh7fNgGs}} provides a robust, fast transcription service for Ghanaian-accented English speech containing mathematical and scientific content. We used OpenAI’s Whisper model \cite{radford2023}  for speech transcription and word error rate (WER) and latency as our evaluation metrics. We previously evaluated all versions of Whisper with a limited sample of 3 audios (approximately 15 seconds long) from the NSMQ competition which consists of speech with Ghanaian accents, along with their corresponding transcripts \cite{boateng2023nsmqai}. The WERs are quite high, making the case to fine-tune the model with Ghanaian accented speech in the future. None of the models attained the best scores for both metrics, which warranted the need for a trade-off between the model metrics. An ideal model would have the lowest WER and latency values. We selected the Medium (English) model to deploy for use as it had a good trade-off between WER and latency (31.61\% and 0.94 seconds respectively).

\subsection{Question Extraction}
The Question Extraction (QE) system extracts relevant portions of the riddles (clues) by inferring the start and end of each riddle and segmenting the clues. To infer the start and end of each riddle, we implemented a check for an exhaustive list of specific phrases that tend to be said at the beginning of each riddle such as “first riddle, second riddle, we begin, etc.” To segment the clues, we fine-tuned a BERT (Tiny) model to classify automatically transcribed text of audio chunks received from STT as clues or non-clues. We used a dataset of 184  manually annotated clues (n=81) and non-clues (n=103) from past riddles. We used 80\% as the train and 20\% as the test set. We had a separate held-out validation set of 173 samples (clues=119, non-clues=54) which we used for the final evaluation. We trained it for 10 epochs using an Adam optimizer (with a learning rate of 5e-5) and a batch size of 8. As a baseline, we extracted (1) TF-IDF and (2) SBERT embeddings as features and trained with various ML models such as random forest, support vector machine (SVM), and logistic regression. The BERT approach performed the best with 97\% and 91\% balanced accuracies for the test set and held-out validation set respectively compared to the best baseline approach of  SBERT + SVM of  97\% and SBERT + logistic regression of 87\% for the test set and heldout validation set respectively. Hence we used the BERT approach for the deployment.

\subsection{Question Answering}
The Question Answering (QA) system \footnote{QA System Demo: \url{https://youtu.be/VfkxZAdZ2PA}} takes as input a riddle that consists of a clue or set of clues, and then attempts to provide answers swiftly and accurately ahead of human contestants. We implemented and evaluated two approaches: Extractive QA and Generative QA. 

The Extractive QA approach involves retrieving contexts relevant to the current set of clues from a semantic search engine such as our custom-built semantic search engine consisting of a custom vector database computed with Sentence-BERT (SBERT) \cite{reimers2019} over passages from the Simple English Wikipedia dataset or Kwame for Science which used SBERT and a science dataset + Simple English Wikipedia \cite{boateng2023}.  We passed the concatenated clues into the semantic search engine and retrieved the top three passages with the highest similarity scores. We then passed the concatenated clues into an extractive QA model (DistilBERT) as “questions”, while the retrieved passages are provided as the “context” to return a span of text as an answer. 

The Generative QA approach uses generative models (Falcon-7b-Instruct and Mistral-7B-Instruct-v0.1) along with a well-developed prompt to generate an answer given a set of riddle clues. We made use of prompt engineering to develop a high-quality prompt that takes the clues as part of the prompt. The prompt (1) asked the model to take on the role of an expert — a science prodigy, (2) used Chain-of-Thought (CoT) prompting by asking it to reason through the clues, (3) stated a penalty for deviating from the instruction to provide a short answer, (4) used few-shots learning by giving an example of a riddle and an answer, and (5) asked for a structured output — JSON. 

Furthermore, we implemented a confidence modeling pipeline that produces an estimate of the confidence of the QA system to decide whether or not to attempt to answer the riddle after a clue has been received or wait for more clues. For a riddle, we pass the first clue as an input to the model, generate three answers to the riddle, and keep a running count of the number of times each answer is generated for subsequent input clues and compare against a threshold value which we determine empirically after evaluating different threshold values. If any answer’s count is equal to or more than the threshold value, we then return that answer as our answering attempt.

\begin{figure}[t]
\centering
\includegraphics[width=\linewidth]{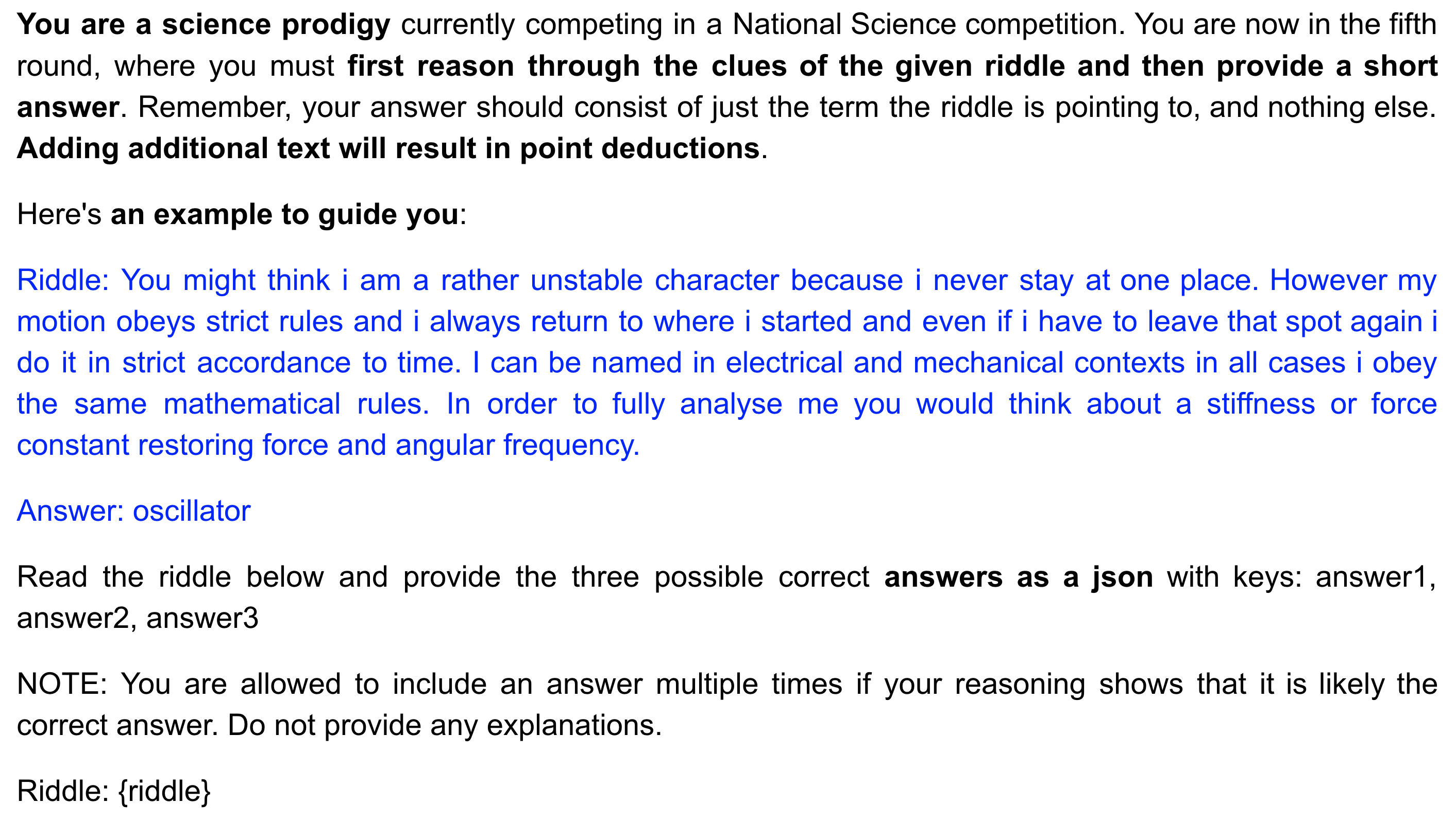}
\caption{Prompt for QA Model} 
\label{fig:qa_prompt}
\end{figure}

We used Exact Match (EM) accuracy as the primary metric and a secondary metric, Fuzzy Match (FM) accuracy. With EM, we search for instances where the generated answer exactly matches the ground truth answer for a given riddle or the alternative answers if there are any. With FM, we perform a more relaxed comparison and check if any of the ground truth answers is a substring of the answer generated by the model. For example, (ground truth: tissue, model answer: tissues, will yield True for fuzzy match). We opt for EM as our primary evaluation metric, given the quiz’s nature, which requires that the answers provided by the competing teams precisely match the term the riddle points to. Additionally, we employ FM to assess overall model performance and to gauge generated answers' proximity to the ground truth.

We perform two sets of evaluations. The first evaluation uses all clues concatenated as input to our models and assesses performance (Table \ref{tab:qa_results_all}). This case assumes that the QA system has all the clues and enables us to compare different models. The second evaluation uses the best approach and model from the first evaluation and integrates the confidence modeling approach to simulate the live deployment offline (Table \ref{tab:qa_results_live}). Given that the live mode uses an automatic transcript of 5 seconds of audio, we estimated how many words are in such a chunk (about 7 words). We break our text of the riddles into those chunks and pass them one after another while appending subsequent chunks as input to the model as it attempts to answer at some point. We compared the performance of our 2 approaches —  Extractive QA (using SBERT + Kwame for Science dataset or Simple Wiki for retrieval and DistilBERT for extraction) and Generative QA (using Mistral-7b-Instruct-v0.1 and Falcon-7b-Instruct) with GPT 3.5 and a human benchmark which we computed using our annotations about the clue number on which the contestanting students answered the riddles.  For all experiments and evaluations, we run on either a Tesla T4 GPU with 16GB VRAM, or a single A100 GPU with 40GB VRAM on Google Colab.

Our results are shown in Tables \ref{tab:qa_results_all} and \ref{tab:qa_results_live}. From  Table \ref{tab:qa_results_all}, our generative approach using Mistral performed better than the extractive approach but worse than GPT 3.5 and students. Hence, we used Mistral for our second evaluation with confidence modeling and deployed it. GPT 3.5 performed better than our approach but worse than the best human contestants for both evaluations 1 and 2 (Table \ref{tab:qa_results_live}). Overall, these show advancements in generative models for solving science questions. However, the proprietary and closed nature of GPT 3.5, underscores the need to create more accurate open alternatives that will be accessible to students and trainers in Africa and other low-resourced environments. Fine-tuning our open-source models in the future could potentially accomplish that.

\begin{table}[ht]
\centering
\caption{Evaluation Using All Clues on 2019 (n=156) and 2020 (n=160) riddles dataset.}
\label{tab:qa_results_all}
\begin{tabular}{|c|cc|cc|}
\hline
\textbf{Model}           & \multicolumn{2}{c|}{\textbf{EM (\%)}}              & \multicolumn{2}{c|}{\textbf{FM (\%)}}              \\ \cline{2-5} 
\multicolumn{1}{|l|}{}   & \multicolumn{1}{c|}{\textbf{2019}} & \textbf{2020} & \multicolumn{1}{c|}{\textbf{2019}} & \textbf{2020} \\ \hline
DistilBERT (Kwame)    & \multicolumn{1}{c|}{0}             & 0             & \multicolumn{1}{c|}{0}             & 0.63          \\ \hline
DistilBERT (Simple Wiki) & \multicolumn{1}{c|}{1.28}          & 0             & \multicolumn{1}{c|}{2.56}          & 1.25          \\ \hline
Falcon                   & \multicolumn{1}{c|}{22.44}         & 14.37         & \multicolumn{1}{c|}{33.33}         & 36.25         \\ \hline
Mistral                  & \multicolumn{1}{c|}{38.46}         & 27.5          & \multicolumn{1}{c|}{54.49}         & 45.62         \\ \hline
GPT 3.5                  & \multicolumn{1}{c|}{40.38}         & 33.75         & \multicolumn{1}{c|}{70.51}         & 68.12         \\ \hline
Students                 & \multicolumn{1}{c|}{76.3}          & 75            & \multicolumn{1}{c|}{N/A}           & N/A           \\ \hline
\end{tabular}
\end{table}

\begin{table}[ht]
\centering
\caption{Mock Live Environment Evaluation for Mistral and ChatGPT on 2019 (n=156) and 2020 (n=160) riddles dataset.}
\label{tab:qa_results_live}
\begin{tabular}{|c|cc|cc|}
\hline
\textbf{Model}         & \multicolumn{2}{c|}{\textbf{EM (\%)}}              & \multicolumn{2}{c|}{\textbf{FM (\%)}}              \\ \cline{2-5} 
\multicolumn{1}{|l|}{} & \multicolumn{1}{c|}{\textbf{2019}} & \textbf{2020} & \multicolumn{1}{c|}{\textbf{2019}} & \textbf{2020} \\ \hline
Mistral                & \multicolumn{1}{c|}{12.82}         & 8.75          & \multicolumn{1}{c|}{21.15}         & 24.38         \\ \hline
GPT 3.5                & \multicolumn{1}{c|}{28.85}         & 23.75         & \multicolumn{1}{c|}{33.97}         & 35.63         \\ \hline
Students             & \multicolumn{1}{c|}{76.3}          & 75.0          & \multicolumn{1}{c|}{N/A}           & N/A           \\ \hline
\end{tabular}
\end{table}

\subsection{Text-to-Speech}
The TTS system synthesizes answers generated by QA  into speech with a Ghanaian accent  \footnote{TTS System Demo: \url{https://youtu.be/KuOxxAk_Qqk}}. We used VITS \cite{kim2021}, an end-to-end model that we fine-tuned using voice samples from three Ghanaian speakers resulting in an output speech similar to those speakers \footnote{TTS Voice Sample: \url{https://youtu.be/dwg7izBMFGA}}. The training datasets featured audio samples along with automatic transcripts using Whisper (which were manually corrected) from three Ghanaians (with their permission), each with unique recordings extracted from various sources: Speaker 1: TEDX talk (20 minutes), Speaker 2: podcast interview (22 minutes), speaker 3: YouTube recording (1 hour). We converted our models to the Open Neural Network Exchange (ONNX) format for deployment. ONNX is a file format that allows for easy integration of ML models across various frameworks like Tensorflow and PyTorch \cite{onnx}. ONNX Runtime optimizes latency, throughput, memory utilization, and binary size and allows users to run ML models efficiently.

We performed two evaluations using the ONNX versions of the models. Evaluation 1 involved synthesizing 30 samples of scientific and mathematical speech from past NSMQ questions and Evaluation 2 involved synthesizing 30 samples of conversational speech. We compared the 3 speaker-specific models. We used Mean Opinion Score (MOS), Word Error Rate (WER), and latency as evaluation metrics. The automatic MOS is an objective evaluation of how ‘natural’ the synthesized speech sounds with a range from 1 (bad) to 5 (excellent) \cite{lo2019}. The WER of the synthesized speech measures the intelligibility of the synthesized speech, whereas the latency measures the inference speed of the model. The results (Table \ref{tab:tts_results}) show that the Speaker 1 model achieves the best WER. The models generally exhibit high MOS scores, indicating a human-like sound. However, they struggle with scientific and mathematical text synthesis as shown by the poor WER. Qualitatively, audios synthesized with these models sound Ghanaian and similar to the original speakers, with a slight robotic undertone and suboptimal intelligibility for single words. We did not use Speaker 1 to avoid confusion in the live deployment since it is the voice of the quiz mistress. Consequently, we deployed Speaker 3 due to its lower latency and WER compared to Speaker 2. 

\begin{table}[ht]
\centering
\caption{Evaluation of TTS Models}
\label{tab:tts_results}
\resizebox{\textwidth}{!}{%
\begin{tabular}{|l|ccc|ccc|}
\hline
\textbf{}      & \multicolumn{3}{c|}{\textbf{Evaluation 1}}                                                                       & \multicolumn{3}{c|}{\textbf{Evaluation 2}}                                                                       \\ \hline
\textbf{Model} & \multicolumn{1}{c|}{\textbf{Mean Latency (s)}} & \multicolumn{1}{c|}{\textbf{Mean WER (\%)}} & \textbf{Mean MOS} & \multicolumn{1}{c|}{\textbf{Mean Latency (s)}} & \multicolumn{1}{c|}{\textbf{Mean WER (\%)}} & \textbf{Mean MOS} \\ \hline
Speaker 1      & \multicolumn{1}{c|}{1.05}                      & \multicolumn{1}{c|}{35.41}                  & 3.00              & \multicolumn{1}{c|}{1.11}                      & \multicolumn{1}{c|}{6.99}                   & 2.81              \\ \hline
Speaker 2      & \multicolumn{1}{c|}{1.28}                      & \multicolumn{1}{c|}{70.45}                  & 3.12              & \multicolumn{1}{c|}{1.24}                      & \multicolumn{1}{c|}{44.17}                  & 3.32              \\ \hline
Speaker 3      & \multicolumn{1}{c|}{1.08}                      & \multicolumn{1}{c|}{63.12}                  & 2.84              & \multicolumn{1}{c|}{1.05}                      & \multicolumn{1}{c|}{17.91}                  & 2.84              \\ \hline
\end{tabular}
}
\end{table}

\section{Real-World Deployment and Evaluation of Brilla AI}
In its debut in the NSMQ Grand Finale in October 2023, Brilla AI answered one of the 4 riddles ahead of the 3 human contesting teams, unofficially placing second (tied) and achieved a 25\% EM accuracy \footnote{Brilla AI Debut: \url{https://youtu.be/2AUpiVB6zA4}} \cite{joynews}. For the first riddle, the QE component could not detect the start of the riddle as STT wrongly transcribed “first riddle” as “test riddle”. Consequently, no clues were extracted, QA received no data, and thus could not answer the riddle (the answer shown was from a previous test attempt). For the second and third riddles, the start and ends were detected and the clues were extracted but QA attempted too early (perhaps due to the confidence threshold being quite low) and got the answer wrong. For the fourth and final riddle, QE detected the start of the riddle, could not extract clue one, but extracted the second clue and third clue, and then QA generated an answer, resulting in Brilla AI answering correctly before one of the students answered correctly also (which he did after all the clues). Overall, we recorded a success rate of 25\% unofficially placing second (tied) as the best-performing team answered 2 riddles correctly (excluding the one the AI answered), one team answered only one and the third answered none. This result is an important milestone, especially given it was our first live deployment and there were a lot of real-world challenges like noise, inaccurate transcripts, undetected clues, and irrelevant data sent along to QA. Our automatic transcripts were not always accurate and some clues were not detected which all compounded and posed challenges for QA. Our real-time transcription began to lag the live stream of the contest because of the sequential processing of our pipeline, which resulted in our AI appearing to have provided an answer after all clues, even though it did so only after receiving the 3rd clue.  This issue could have resulted in a late answer.

\section{Challenges, Limitations, and Future Work}
One key challenge we have is curating the NSQM data. At the end of each contest, the competing schools and their respective scores are shared via images or PDFs on social media and blogs. Sometimes finding these materials is difficult as they may be scattered on the web or may not have been posted by the media channel in charge. Video recording errors due to power surges and Internet connection problems also make getting accurate data difficult. 

In the web app, the sequential approach of calling APIs was a bottleneck which caused some latency issues for the live deployment. We plan to explore parallelizing operations to remove such bottlenecks from the system. Also, we plan to deploy the web app and ML models on Google Cloud Platform to ensure that both the front and backends of Brilla AI are running on cloud systems to increase the security and stability of the web app. 

We plan to fine-tune our STT model since we did not do so in this work due to a lack of complete NSMQ annotated data, which could improve performance. To improve the QA performance, we will explore using RAG after completing the curation and processing of the science textbooks. We will also explore using reinforcement learning to improve our confidence modeling approach. For TTS, we will curate a dataset that includes recordings of scientific and mathematical expressions, as well as single-word answers, to address its shortcomings.

\section{Conclusion}
In this work, we built Brilla AI, an AI contestant that we deployed to unofficially compete remotely and live in the Riddles round of the 2023 NSMQ Grand Finale, the first of its kind in the 30-year history of the competition. Our AI answered one of the 4 riddles ahead of the 3 human contesting teams, unofficially placing second (tied). Our next step is to improve the performance of our system ahead of 2024 NSMQ during which we plan to unofficially compete in both Round 5 (Riddles) and Round 4 (True or False) for all 5 stages of the competition. Aside from being an interesting intellectual challenge, we are building an AI that addresses the unique context of Africans — transcribes speech with a Ghanaian accent, provides answers to scientific questions drawn from a Ghanaian quiz, and says answers with a Ghanaian accent — that could be integrated into an education tool. Imagine a student in a rural part of Ghana calling a toll-free number with a feature phone, asking this AI numerous science questions, and the AI understands her even with her Ghanaian accent and then provides explanations using local context in a Ghanaian accent. That vision is our long-term goal and would radically transform learning support for students across Africa, enabling millions of young people to have one-on-one interactions and support even with limited access to teachers, computers, or even smartphones leading to the democratizing of science education across Africa!